\documentclass[10pt,twocolumn,letterpaper]{article}

\usepackage{style/cvpr}
\usepackage{times}
\usepackage{epsfig}
\usepackage{gensymb}
\usepackage{graphicx}
\usepackage{amsmath}
\usepackage{amssymb}
\usepackage{caption}
\usepackage{color}
\usepackage{siunitx}
\usepackage{multirow}
\usepackage{multicol}
\usepackage{wrapfig}
\usepackage{subcaption}
\usepackage{etoolbox}
\usepackage{xcolor}
\usepackage{booktabs}
\usepackage{cuted}

\usepackage[pagebackref=true,breaklinks=true,letterpaper=true,colorlinks,bookmarks=false]{hyperref}

\cvprfinalcopy 

\def\cvprPaperID{9648} 

\ifcvprfinal\fi
\begin{document}

\title{Streaming Object Detection for 3-D Point Clouds}

\author{
  {\bf Wei Han}$^{\dagger}$, {\bf Zhengdong Zhang} $^{\dagger}$, {\bf Benjamin Caine}$^{\dagger}$, {\bf Brandon Yang}$^{\dagger}$,
  {\bf Christoph Sprunk}$^{\ddagger}$,\\ {\bf Ouais Alsharif}$^{\ddagger}$, {\bf Jiquan Ngiam}$^{\dagger}$, {\bf Vijay Vasudevan}$^{\dagger}$,
  {\bf Jonathon Shlens}$^{\dagger}$, {\bf Zhifeng Chen}$^{\dagger}$\\
  $^{\dagger}$Google Brain, $^{\ddagger}$Waymo\\
  \texttt{\{weihan,zhangzd\}@google.com}\\
}


\maketitle

\begin{abstract}
Autonomous vehicles operate in a dynamic environment, where the speed with which a vehicle can perceive and react impacts the safety and efficacy of the system.
LiDAR provides a 
prominent sensory modality that informs many existing perceptual systems including object detection, segmentation, motion estimation, and action recognition.
The latency for perceptual systems based on point cloud data can be dominated by the amount of time for a complete rotational scan (e.g. 100 ms). 
This built-in data capture latency is artificial, and based on treating the point cloud as a camera image in order to leverage camera-inspired architectures.
However, unlike camera sensors, most LiDAR point cloud data is natively a {\it streaming} data source in which laser reflections are sequentially recorded based on the precession of the laser beam. 
In this work, we explore how to build an object detector that removes this artificial latency constraint, and instead operates on native streaming data in order to significantly reduce latency.
This approach has the added benefit of reducing the peak computational burden on inference hardware by spreading the computation
over the acquisition time for a scan.
We demonstrate a family of streaming detection systems based on sequential modeling through a series of modifications to the traditional detection meta-architecture.
We highlight how this model may achieve competitive if not superior predictive performance with state-of-the-art, traditional non-streaming detection systems while achieving significant latency gains (e.g. $1/15^\text{th}-1/3^\text{rd}$ of peak latency).
Our results show that operating on LiDAR data in its native streaming formulation offers several advantages for self driving object detection -- advantages that we hope will be useful for any LiDAR perception system where minimizing latency is critical for safe and efficient operation.
\end{abstract}
\section{Introduction}
\begin{figure}[t]
\centering

\includegraphics[width=7.5cm]{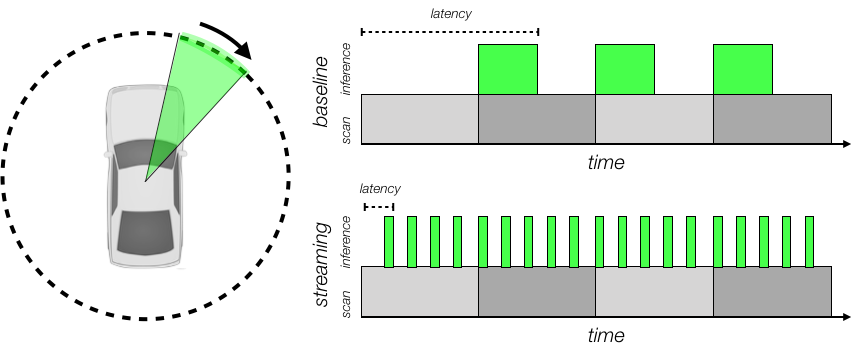}
\vspace{0.2cm}

\setlength{\tabcolsep}{4pt}
{\small
\bgroup
\def\arraystretch{1.1}
\begin{tabular*}{7.2cm}{c|c|cccc}
     meta-architecture & baseline & \multicolumn{4}{c}{streaming} \\
     \hline
     localized RF & & \checkmark & \checkmark  & \checkmark & \checkmark\\
     stateful NMS & & & \checkmark  & \checkmark & \checkmark \\
     stateful RNN & & & & \checkmark & \checkmark \\
     larger model & & & & & \checkmark \\
     \hline
     {\it accuracy (mAP)} &  &  &  & & \\
     pedestrians & 54.9 & 40.1 & 52.9 & 53.5 & 60.1 \\
     vehicles & 51.0 & 10.5 & 39.2 & 48.9 & 51.0 \\

 \end{tabular*}
\egroup
}
\caption{\textbf{Streaming object detection pipelines computation to minimize latency without sacrificing accuracy}. LiDAR accrues a point cloud incrementally based on a rotation around the $z$ axis.
Instead of artificially waiting for a complete point cloud scene based on a \ang{360} rotation ({\it baseline}), we perform inference on subsets of the rotation to pipeline computation ({\it streaming}). Gray boxes indicate the duration for a complete rotation of a LiDAR (e.g. \mbox{100 ms} \cite{waymo-open-dataset,geiger2013vision}). Green boxes denote inference time. The expected latency for detection -- defined as the time between a measurement and a detection decreases substantially in a streaming architecture (dashed line). At 100 ms scan time, the expected latency reduces from $\sim 120$ ms (baseline) versus $\sim 30$ ms (streaming), i.e. $>$3$\times$ (see text for details). The table compares the detection accuracy (mAP) for the baseline on pedestrians and vehicles to several streaming variants \cite{lang2018pointpillars}.}
\label{fig:intro}
\end{figure}

Autonomous driving systems require detection and localization of objects to effectively respond to a dynamic environment \cite{geiger2013vision,nuscenes2019}.
As a result, self-driving cars (SDCs) are equipped with an array of sensors to robustly identify objects across highly variable environmental conditions \cite{cho2014multi,thrun2006stanley}.
In turn, driving in the real world requires responding to this large array of
data with minimal latency to maximize the opportunity for safe and effective navigation \cite{kim2013parallel}.

LiDAR represents one of the most prominent sensory modalities in SDC systems \cite{cho2014multi,thrun2006stanley} informing object detection \cite{yang2018pixor,zhou2018voxelnet,yan2018second,yang2018hdnet}, region segmentation \cite{lindner2009multi,lim2017implementation} and motion estimation \cite{jeon2018lidar,zhang2015visual}.
Existing approaches to LiDAR-based perception derive from a family of camera-based approaches \cite{ren2015faster,liu2016ssd,he2017mask,chen2017deeplab,ilg2017flownet,dosovitskiy2015flownet}, requiring a complete 360$\degree$ scan of the environment. This artificial requirement to have the complete scan limits the minimum latency a perception system can achieve, and effectively inserts the LiDAR scan period into the latency \footnote{
LiDAR typically operates with a 5-20 Hz scan rate. We focus on 10 Hz (i.e. 100 ms period), because several prominent academic datasets employ this scan rate \cite{waymo-open-dataset,geiger2013vision}, however our results may be equally applied across the range of scan rates available.}. Unlike CCD cameras, many LiDAR systems are {\it streaming} data sources, where data arrives sequentially as the laser rotates around the $z$ axis \cite{ackerman2016lidar,hecht2018lidar}.

Object detection in LiDAR-based perception systems \cite{yang2018pixor,zhou2018voxelnet,yan2018second,yang2018hdnet} presents a unique and important opportunity for re-imagining LiDAR-based meta-architectures in order to significantly minimize latency.
Previous work in camera-based detection systems reduced latency $\sim 2\times$ by introducing a ``single-stage'' meta-architecture \cite{liu2016ssd,redmon2016you,lin2017focal}, however the resulting systems typically suffer from a notable drop in detection accuracy \cite{huang2017speed} (but see \cite{lin2017focal,law2018cornernet}).
LiDAR-based perception systems offer an opportunity for significantly improving latency, but without detrimenting detection accuracy by leveraging the streaming nature of the data.
In particular, streaming LiDAR data permits the design of meta-architectures which operate on the data as it arrives, in order to pipeline the sensory readout with the inference computation to significantly reduce latency (Figure \ref{fig:intro}).

In this work, we propose a series of modifications to standard meta-architectures that may generically adapt an object detection system to operate in a streaming manner.
This approach combines traditional elements of single-stage detection systems \cite{redmon2016you,liu2016ssd} as well as design elements of sequence-based learning systems \cite{chiu2018state,jaitly2015neural,graves2012sequence}. The goal of this work is to show how we can modify an existing object detection system -- with a minimum set of changes and the addition of new operations -- to efficiently and accurately emit detections as data arrives (Figure \ref{fig:intro}). We find that this approach matches or exceeds the performance of several baseline systems \cite{lang2018pointpillars,ngiam2019starnet}, while substantially reducing latency.
For instance, a family of streaming models on pedestrian detection achieves up to 60.1 mAP compared to 54.9 mAP for a baseline non-streaming model, while reducing the expected latency  $> 3\times$.
In addition, the resulting model better utilizes computational resources by pipelining the computation throughout the duration of a LiDAR scan.
We demonstrate through this work that designing architectures to leverage the native format of LiDAR data achieves substantial latency gains for perception systems generically that may improve the safety and efficacy of SDC systems.

\section{Related Work}

\subsection{Object detection in camera images}

Object detection has a long history in computer vision as a central task in the field. Early work focused on framing the problem as a two-step process consisting of an initial search phase followed by a final discrimination of object location and identity \cite{felzenszwalb2010object,dean2013fast,uijlings2013selective}. Such strategies proved effective for academic datasets based on camera imagery \cite{everingham2010pascal,lin2014microsoft}.

The re-emergence of convolutional neural networks (CNN) for computer vision \cite{krizhevsky2012imagenet,krizhevsky2009learning} inspired the field to harness both the rich image features and final training objective of a CNN model for object detection \cite{sermanet2013overfeat}. In particular, the features of a CNN trained on an image classification task proved sufficient for providing reasonable candidate locations for objects \cite{girshick2014rich}. Subsequent work demonstrated that a single CNN may be trained in an end-to-end fashion to sub-serve for both stages of an object detection system \cite{ren2015faster,girshick2015fast}. The resulting two-stage systems, however, suffered from relatively poor computational performance, as the second stage necessitated performing inference on all candidate locations leading to trade-offs between thoroughly sampling the scene for candidate locations and predictive performance for localizing objects \cite{huang2017speed}.

The computational demands of a two-stage systems paired with the complexity of training such a system motivated researchers to consider one-stage object detection, where by a single inference pass in a CNN suffices for localizing objects \cite{liu2016ssd,redmon2016you}. One-stage object detection systems lead to favorable computational demands at the sacrifice of predictive performance \cite{huang2017speed} (but see \cite{lin2017focal,lin2016feature} for subsequent progress). 

\subsection{Object detection in videos}

Object detection for videos reflects possibly the closest set of methods relevant to our proposed work. In video object detection, the goal is to detect and track the one or more objects over subsequent camera image frames. Strategies for tackling this problem include breaking up the problem into a computationally-heavy detection phase and a computationally-light tracking phase \cite{liu2018mobile}, and building blended CNN recurrent architectures for providing memory between time steps of each frame \cite{mcintosh2018recurrent,liu2019looking}. 

Recent methods have also explored the potential to persist and update a memory of the scene. Glimpses of a scene are provided to the model; for example, views from specific perspectives \cite{henriques2018mapnet} or cropped regions \cite{chai2019patchwork}, and the model is required to piece together the pieces to make predictions. 

In our work, we examine the possibility of dividing a \textit{single frame} into slices that can be processed in a streaming fashion. A time step in our setup correspond to a slice of one frame. An object may appear across multiple slices, but generally, each slice contains distinct objects. This may require similar architectures to video object detection (e.g., convolutional LSTMs) in order to provide a memory and state of earlier slices for refining or merging detections \cite{xingjian2015convolutional,pinheiro2014recurrent,wu2016google}.

\subsection{Object detection in point clouds}

The prominence of LiDAR systems in self-driving cars necessitated the application of object detection to point cloud data \cite{geiger2013vision,nuscenes2019}.
Much work has employed object detection systems originally designed for camera images to point cloud data by projecting such data from a Bird's Eye View (BEV) \cite{yang2018pixor,luo2018fast,yang2018hdnet,lang2018pointpillars} (but see \cite{meyer2019lasernet}) or a 3-D voxel grid  \cite{zhou2018voxelnet,yan2018second}.
Alternatively, some methods have re-purposed two stage object detector design with a region-proposal stage but replacing the feature extraction operations \cite{yang2018ipod,shi2019pointrcnn,qi2018frustum}. 

In parallel, others have pursued replacing a discretization operation with a  featurization based on native point-cloud data \cite{qi2017pointnet,qi2017pointnet++}. Such methods have led to methods for building detection systems that blend aspects of point cloud featurization and traditional object detectors on cameras \cite{lang2018pointpillars,zhou2019multiview,ngiam2019starnet} to achieve favorable performance for a given computational budget.

\section{Methods}

\begin{figure}[t]
\centering
\includegraphics[width=0.85\linewidth]{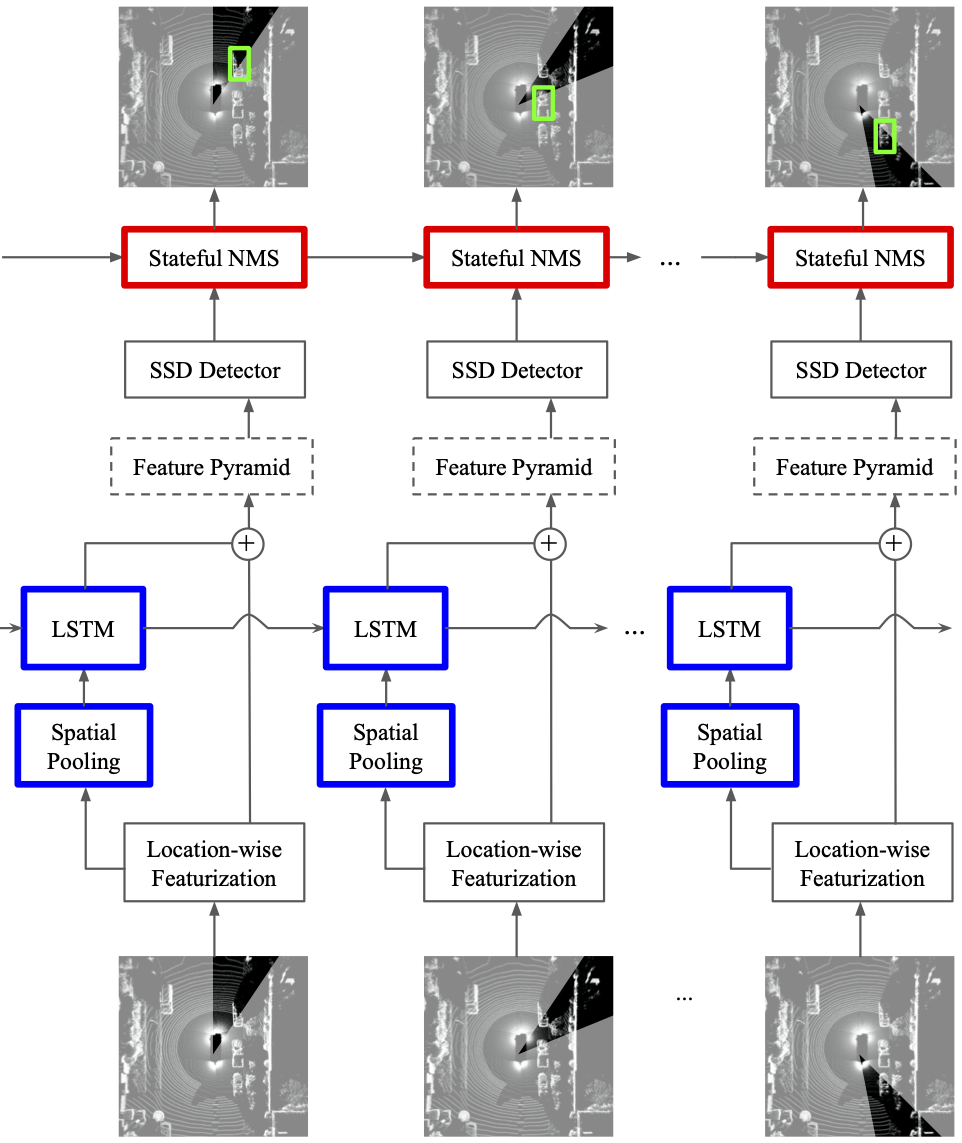}
\caption{\textbf{Diagram of streaming detection architecture.} A streaming object detection system processes a spatially restricted slice of the scene. We introduce two stateful components: Stateful NMS (red) and a LSTM (blue) between input slices. Detections produced by the model are denoted in green boxes. Dashed line denotes feature pyramid uniquely employed in \cite{lang2018pointpillars}.}
\label{fig:architecture}
\end{figure}


\subsection{Streaming LiDAR inputs}
\label{sec:streaming-lidar-inputs}

A LiDAR system for an SDC measures the distance to objects by shining multiple lasers at fixed inclinations and measuring the reflectance in a sensor \cite{cho2014multi,thrun2006stanley}. The lasers precess around the $z$ axis, and make a complete rotation at a 5-20 Hz scan rate. Typically, SDC perception systems artificially wait for a complete 360$\degree$ rotation before processing the data.

In this work, we simulate a streaming system with the Waymo Open Dataset \cite{waymo-open-dataset} by artificially manipulating the point cloud data \footnote{KITTI \cite{geiger2013vision} is the most popular LiDAR detection dataset, however this dataset provides
annotations within a 90$\degree$ frustum. The Waymo Open Dataset provides a completely annotated 360$\degree$ point cloud which is necessary to demonstrate the efficacy of the streaming architecture across all angular orientations.}.
The native format of point cloud data are {\it range images} whose resolution in height and width correspond to the number of lasers and the rotation speed and laser pulse rate \cite{meyer2019lasernet}.
In this work, we artificially slice the input range image into $n$ vertical strips along the image width to provide an experimental setup to experiment with streaming detection models.

\subsection{Streaming object detection}

This work introduces a meta-architecture for adapting object detection systems for point clouds to operate in a streaming fashion. We employ two models as a baseline to demonstrate how the proposed changes to the meta-architecture are generic, and may be employed in notably different detection systems.

We first investigate PointPillars \cite{lang2018pointpillars} as a baseline model because it provides competitive performance in terms of predictive accuracy and computational budget.
The model divides the $x$-$y$ space into a top-down 2D grid, where each grid cell is referred to as a {\it pillar}. The points within each non-zero pillar are featurized using a variant of a multi-layer perceptron architecture designed for point cloud data \cite{qi2017pointnet,qi2017pointnet++}. The resulting $d$-dimensional point cloud features are scattered to the grid and a standard multi-scale, convolutional feature pyramid \cite{lin2016feature,zhou2018voxelnet} is computed on the spatially-arranged point cloud features to result in a global activation map. The second model investigated is StarNet \cite{ngiam2019starnet}. StarNet is an entirely point-based detection system which uses sampling instead of a learned region proposal to operate on targeted regions of a point cloud. StarNet avoids usage of global information, but instead targets the computational demand to regions of interest, resulting in a locally targeted activation map. See the Appendix for architecture details for both models.
For both PointPillars and StarNet, the resulting activation map is regressed on to a 7-dimensional target parameterizing the 3D bounding box as well as a classification logit \cite{yan2018second}.
Ground truth labels are assigned to individual anchors based on intersection-over-union (IoU) overlap~\cite{yan2018second,lang2018pointpillars}. To generate the final predictions, we employ oriented, 3-D multi-class
non-maximal suppression (NMS) \cite{girshick2014rich}.

In the streaming object detection setting, models are limited to a restricted view of the scene. We carve up the scene into $n$ slices (Section \ref{sec:streaming-lidar-inputs}) and only supply an individual slice to the model (Figure \ref{fig:architecture}). We simplify the parameterization by requiring that slices are non-overlapping (i.e., the stride between slices matches the slice width). We explore a range of $n$ in the subsequent experiments.

For the PointPillars convolutional backbone, we assume that sparse operators are employed in the convolutional backbone to avoid computation on empty pillars \cite{sparseconv}. Note that no such implementation is required for StarNet because the model is designed to only operate on populate regions of the point cloud \cite{ngiam2019starnet}.

\subsection{Stateful non-maximum suppression}
\label{sec:nms-methods}

Objects which subtend large angles of the LiDAR scan present unique challenges to a steaming detection system which necessarily have a limited range of sensor input (Table \ref{table:error-analysis}).
Hence, we explore a modified NMS technique that maintains \textit{state}. 
Generically, NMS with state may take into account detections in the previous $k$ slices to determine if a new detection is indeed unique.
Therefore, detections from the current slice can be suppressed by those in previous slices. Stateful NMS does not require a complete LiDAR rotation and may likewise operate in a streaming fashion.
In our experiments, for a broad range of $n$, we found that $k=1$ achieves as good of performance as $k$=$n-1$ which would correspond to a global NMS available to a non-streaming system. We explore the selection of $k$ in Section \ref{sec:nms_results}.

\subsection{Adding state with recurrent architectures}

Given a restricted view of the point cloud scene, streaming models can be limited by the context available to make predictions. To increase the amount of context that the model has access to, we consider augmenting the baseline model to maintain a recurrent memory across consecutive slices. This memory may be placed in the intermediate representations of the network.

We select a standard single layer LSTM as our recurrent architecture, although any other RNN architecture may suffice \cite{hochreiter1997long}. 
The LSTM is inserted after the final global representation from either baseline model before regressing on to the detection targets.
For instance, in the PointPillars baseline model, this corresponds to inserting the LSTM after the convolutional representation.
The memory input is then the spatial average pooling for all activations of the final convolutional layer before the feature pyramid \footnote{The final convolutional layer corresponds to the third convolutional layer in PointPillars \cite{lang2018pointpillars} after three strided convolutions. When computing the feature pyramid, note that the output of the LSTM is only added the corresponding outputs of the third convolutional layer.}. Note that the LSTM memory does not hard code the spatial location of the slice it is processing.

Based on earlier preliminary experiments, we employed LSTM with $128$ hidden dimensions and $256$ output dimensions. 
The output of the LSTM is summed with the final activation map by broadcasting across all spatial dimensions in the hidden representation.
More sophisticated elaborations are possible, but are not explored in this work \cite{liu2018mobile,mcintosh2018recurrent,liu2019looking,xingjian2015convolutional,pinheiro2014recurrent,wu2016google}.

\section{Results}

We present all results on the Waymo Open Dataset \cite{waymo-open-dataset}. All models are trained with Adam \cite{kingma2014adam} using the Lingvo machine learning framework \cite{shen2019lingvo} built on top of TensorFlow \cite{abadi2016tensorflow}
\footnote{Code available at \tt{http://github.com/tensorflow/lingvo}}.
We perform hyper-parameter tuning through cross-validated studies and final evaluations on the corresponding test datasets.
In our experiments, we explore how the proposed meta-architecture for streaming 3-D object detection compares to a standard object detection system, i.e. PointPillars \cite{lang2018pointpillars} and StarNet \cite{ngiam2019starnet}. All experiments use the first return from the medium range LiDAR (labeled TOP) in the Waymo Open Dataset, ignoring the four short range LiDARs for simplicity. This results in slightly lower baseline and final accuracy as compared to previous results \cite{ngiam2019starnet,waymo-open-dataset,zhou2019multiview}.

We begin with a simple, generic modification to the baseline architecture by limiting its spatial view to a single slice. This modification permits the model to operate in a streaming fashion, but suffers from a severe performance degradation. We address these issues through stateful variants of non-maximum suppression (NMS) and recurrent architectures (RNN). We demonstrate that the resulting streaming meta-architecture restores competitive performance with favorable computation and latency benefits.
Importantly, such a meta-architecture may be applied generically to most other point cloud detection models~\cite{ngiam2019starnet,zhou2018voxelnet,yan2018second,yang2018pixor,luo2018fast,yang2018hdnet,meyer2019lasernet}.

\subsection{Spatially localized detection degrades baseline models}
\label{sec:spatially-localized}

\begin{figure*}[t]
\begin{center}
{\bf PointPillars} \\
\includegraphics[width=0.39\linewidth]{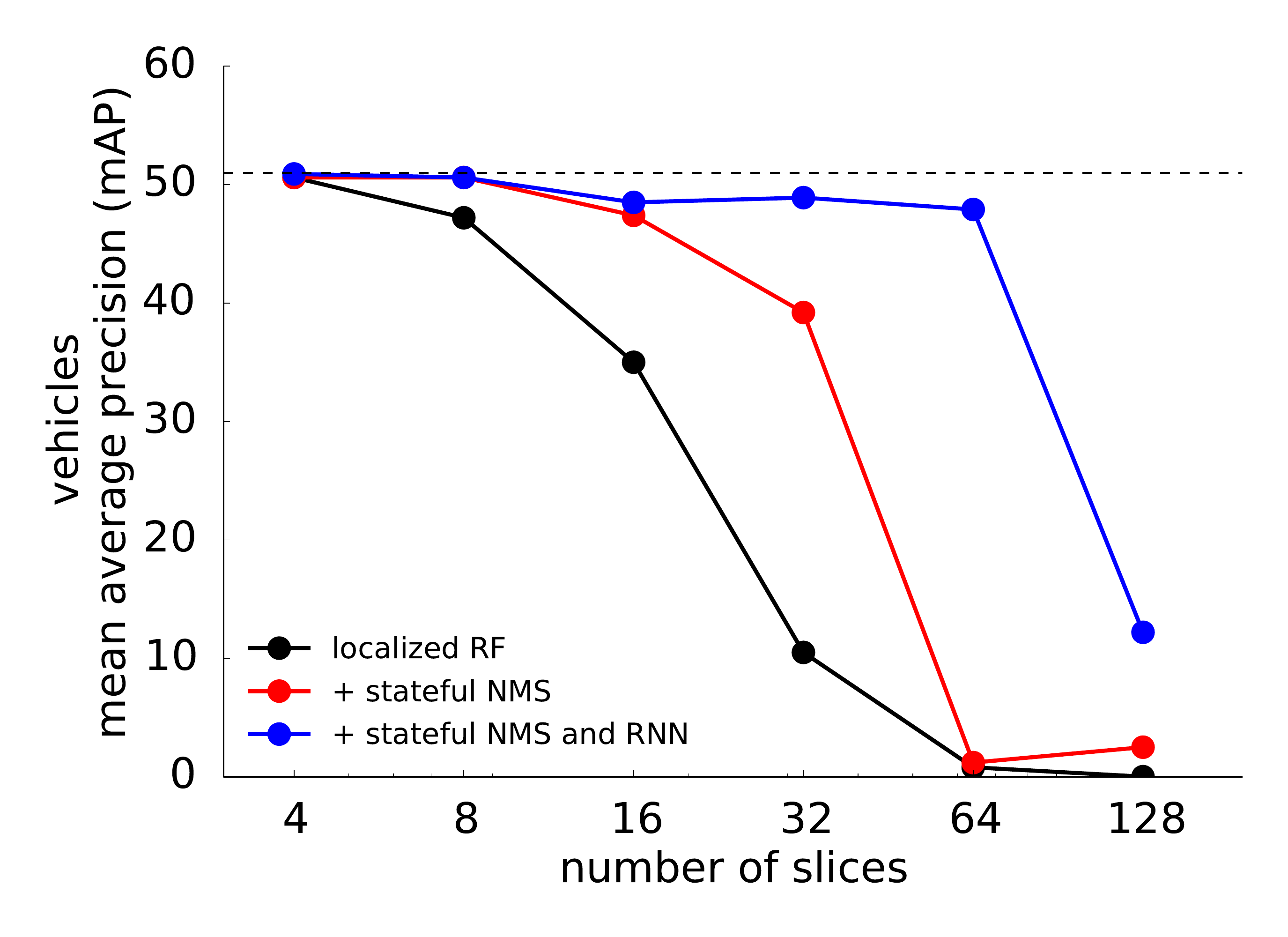}
\includegraphics[width=0.39\linewidth]{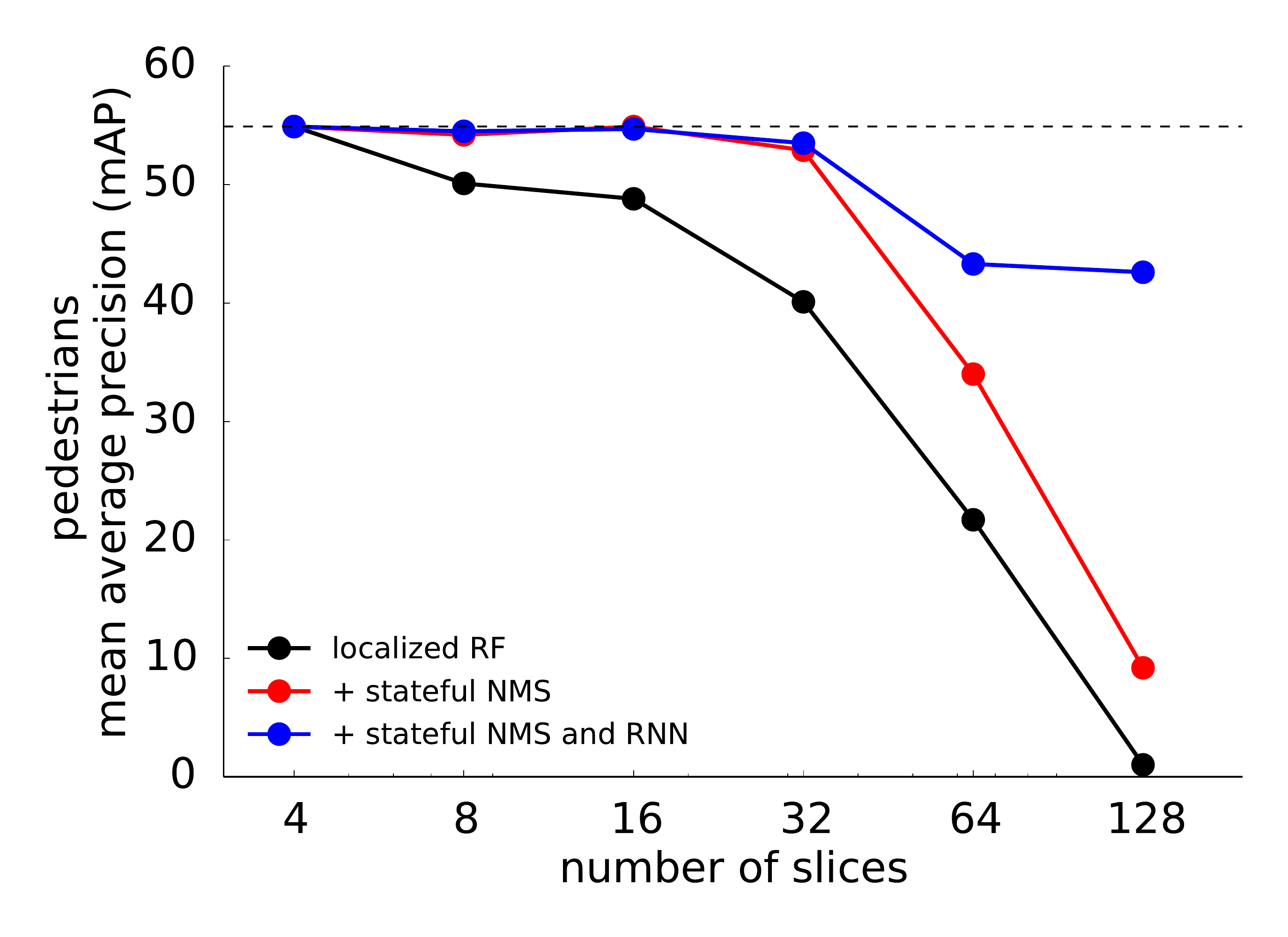} \\
{\bf StarNet} \\
\includegraphics[width=0.39\linewidth]{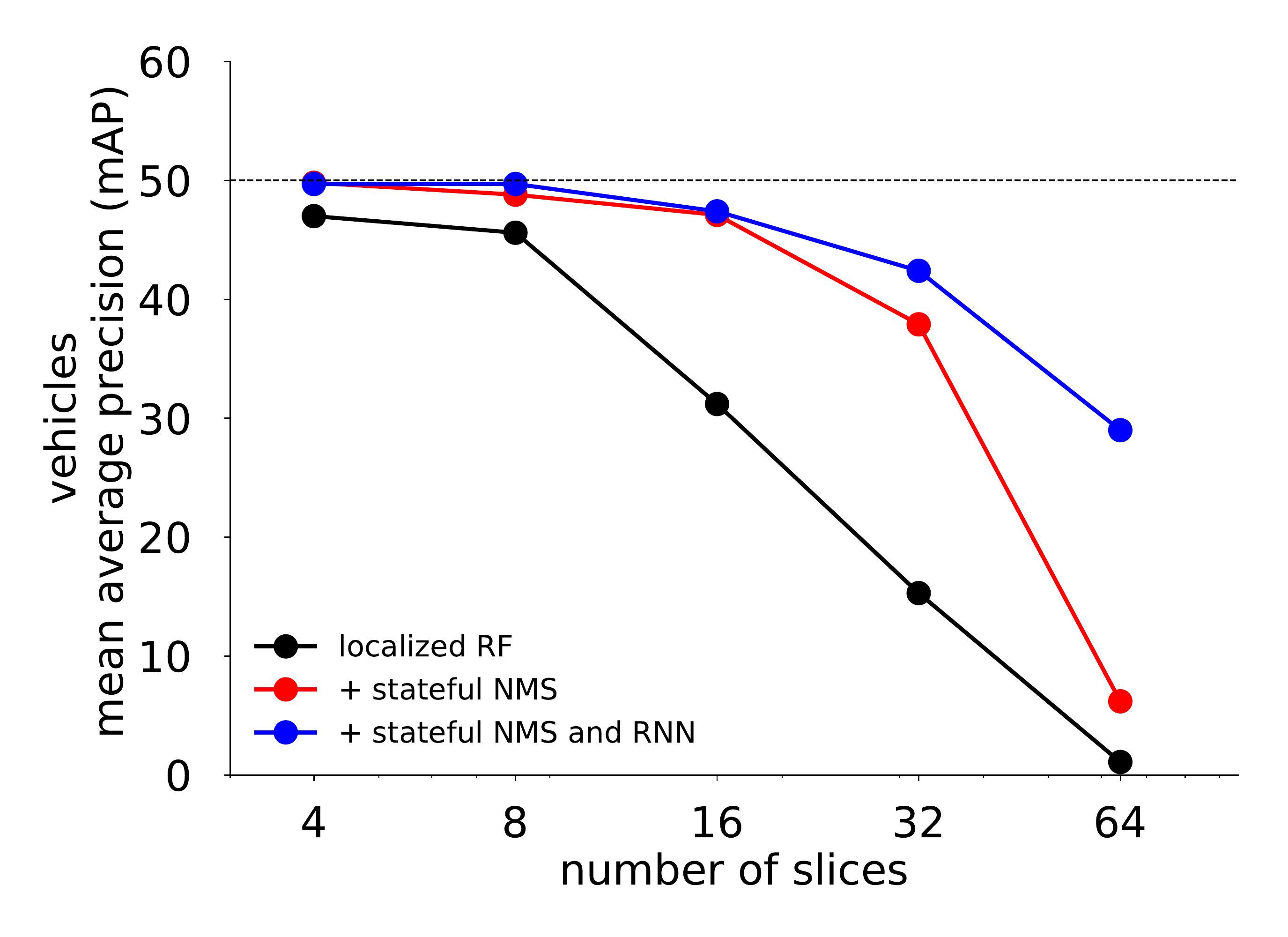}
\includegraphics[width=0.39\linewidth]{figures/Streaming-Detection-AP-Car-starnet.pdf}
\end{center}
\vspace{-0.7cm}
\caption{\textbf{Streaming object detection may achieve comparable performance to a non-streaming baseline.} Mean average precision (mAP) versus the number of slices $n$ within a single rotation for (left) vehicles and (right) pedestrians for (top) PointPillars \cite{lang2018pointpillars} and (bottom) StarNet \cite{ngiam2019starnet}. The solid black line (localized rf) corresponds to the modified baseline that operates on a spatially restricted region. Dashed lines corresponds to baseline models that processes the entire laser spin (vehicle = 51.0\%; pedestrian = 54.9\%). Each curve represents a streaming architecture (see text).}
\label{fig:accuracy-vs-slices}
\end{figure*}

As a first step in building a streaming object detection system, we modify a baseline model to operate on a spatially restricted region of the point cloud. Again, we emphasize that these changes are generic, and may be introduced to most object detectors for LiDAR. 
Such a model trains on a restricted view of the LiDAR subtending an angle in the $x$-$y$ plane (Figure \ref{fig:architecture}).
By operating on a spatially restricted region of data, the model may in principle operate with a reduced latency since inference can proceed before a complete LiDAR rotation (Figure \ref{fig:intro}).

Two free parameters that govern a localized receptive field are the angle of restriction and the stride between subsequent inference operations. To simplify this analysis, we parameterize the angle based on the number of slices $n$ where the angular width is $\ang{360} / n$. We specify the stride to match the angle such that each inference performs inference on non-overlapping slices; overlapping slices is an option, but requires recomputing features on the same points.
We compare the performance of the streaming models against the baselines in Figure \ref{fig:accuracy-vs-slices}.

As the number of slices $n$ grows, the streaming model receives a decreasing fraction of the scene and the predictive performance monotonically decreases (black solid line). This reduction in accuracy may be expected because less sensory data is available for each inference operation.  
All objects sizes appear to be severely degraded by the localized receptive field, although there is a slight negative trend for increasing sizes (Table \ref{table:error-analysis}).
The negative trend is consistent with the observation that vehicle mAP drops off faster than pedestrians, as vehicles are larger and will more frequently cross slices boundaries.
For instance, at $n$=16, the mean average precision is 35.0\% and 48.8\% for vehicles and pedestrians, respectively.
A large number of slices $n$=128 severely degrades model performance for both types.

\begin{table}[!t]
\begin{center}
\bgroup
\def\arraystretch{1.1}

{\small
\begin{tabular}{r|c|c|c|c|c}
& 0-5$\degree$ & 5-15$\degree$ & 15-25$\degree$ & 25-35$\degree$ & $>$35$\degree$ \\ \hline
baseline & 12.6 & 30.9 & 47.2 & 69.1 & 83.4 \\ \hline
\multirow{2}{*}{localized RF} & \,\,\,2.6 & \,\,\,6.6 & \,\,\,9.2 & 13.1 & 14.0 \\
 & \textcolor{red}{-79\%} & \textcolor{red}{-79\%} & \textcolor{red}{-80\%} &  \textcolor{red}{-81\%} & \textcolor{red}{-83\%} \\ \hline
 
\multirow{2}{*}{+stateful NMS} & \,\,\,9.0 & 24.1 & 36.0 & 54.9 & 68.0 \\
 & \textcolor{red}{-28\%} & \textcolor{red}{-22\%} & \textcolor{red}{-23\%} &  \textcolor{red}{-20\%} & \textcolor{red}{-18\%} \\ \hline

+stateful NMS & 10.8 & 29.9 & 46.0 & 65.3 & 82.1 \\
and RNN& \textcolor{red}{-14\%} & \,\,\,\textcolor{red}{-3\%} & \,\,\,\textcolor{red}{-3\%} & \,\,\,\textcolor{red}{-6\%} & \,\,\,\textcolor{red}{-2\%} \\

\end{tabular}
}
\egroup
\end{center}
\caption{{\bf Localized receptive field leads to duplicate detections.} Vehicle detection performance (mAP) across subtended angle of ground truth objects for localized receptive field and stateful NMS ($n$=32 slices) for \cite{lang2018pointpillars}. Red indicates the percent drop from baseline.}
\label{table:error-analysis}
\end{table}

Beyond only observing fewer points with smaller slices, we observe that even at a modest 8 slices the mAP drops compared to the baseline for both pedestrians and vehicles.  Our hypothesis is that when NMS only operates per slice, there are false positives or false negatives created on any object that crosses a border of two consecutive slices.  Because a partial view of the object leads to a detection in slice $k$, another partial view in slice $k+1$ may lead to duplicate detections for the same object. Either one or both of the detections may be low quality or suppressed, leading to a reduction in mAP.  As a result, we next turn to investigating ways in which we can improve NMS in the intra-frame scenario to restore performance compared to the baseline while retaining the streaming model's latency benefits.

\subsection{Adding state to non-maximum suppression boosts performance}
\label{sec:nms_results}

The lack of state in a spatially localized model across slices impedes predictive performance. In subsequent sections, we consider several options for offering state with changes to the meta-architecture.
We first consider revisiting the non-maximum suppression (NMS) operation. NMS provides a standard heuristic method for improving precision by removing highly overlapping predictions \cite{felzenszwalb2010object,girshick2014rich}. With a spatially localized receptive field, NMS has no ability to suppress overlapping detections arising from distinct slices of the LiDAR spin.

\begin{table}[t]
\begin{center}
\bgroup
\def\arraystretch{1.1}
\begin{tabular}{c|c|c|c|c}
\multicolumn{2}{c|}{slices} & localized & global & stateful \\ \hline
\multirow{2}{*}{16} & car & 35.0 & 47.5 & 47.4 \\
 & ped & 48.8 & 54.8 & 54.9 \\ \hline
\multirow{2}{*}{32} & car & 10.5 & 39.2 & 39.0 \\
& ped & 40.1 & 53.1 & 52.9 \\
\end{tabular}
\egroup
\end{center}
\caption{{\bf Stateful NMS achieves comparable performance gains as global NMS.} Table entries report the mAP for detection on vehicles (car) and pedestrians (ped) \cite{lang2018pointpillars}. Localized indicates the mAP for a spatially restricted receptive field. Global NMS boosts performance significantly but is a non-streaming heuristic. Stateful NMS achieves comparable results but is amenable to a streaming architecture.}
\label{table:nms-variants}
\end{table}

We can verify this failure mode by modifying NMS to operate over the concatenation of all detections across {\it all} $n$ slices in a complete rotation of the LiDAR. We term this operation {\it global} NMS. (Note that global NMS does not enable a streaming detection system because a complete LiDAR rotation is required to finish inference.) We compute the detection accuracy for global NMS as a point of reference to measure how many of the failures of a spatially localized receptive field can be rescued. Indeed, Table \ref{table:nms-variants} (localized vs. global) indicates that applying global NMS improves predictive performance significantly.

We wish to develop a new form of NMS that achieves the same performance as global NMS but may operate in a streaming fashion. We construct a simple form of {\it stateful} NMS that stores detections from the previous slice of the LiDAR scene and use the previous slice's detections to rule out overlapping detections (Section \ref{sec:nms-methods}). Stateful NMS does not require a complete LiDAR rotation and may operate in a streaming fashion. Table \ref{table:nms-variants} (global vs. stateful) indicates that stateful NMS provides predictive performance that is comparable to a global NMS operation within $\pm 0.1$ mAP. Indeed, stateful NMS boosts performance of the spatially restricted receptive field across all ranges of slices for both baseline models (Figure \ref{fig:accuracy-vs-slices}, red curve). This observation is also consistent with the fact that a large fraction of the failures across object size are largely systematically recovered (Table \ref{table:error-analysis}), suggesting that indeed duplicate detections across boundaries hamper model performance. These results suggest that this generic change to introduce state into the NMS heuristic may recover much of the performance drop due to a spatially localized receptive field.

\subsection{Adding a learned recurrent model further boosts performance}
Adding stateful NMS substantially improves streaming detection performance, however, the network backbone and learned components of the system only operate on the current slice of the scene and the outputs of the previous slice.

In this section, we examine how a recurrent model may improve predictive performance by providing access to (1) more than just the previous LiDAR slice, and (2) the lower-level features of the model, instead of just the outputs.
In particular, we provide an alteration to the meta-architecture by adding a recurrent layer to the global representation of the network featurization (Figure \ref{fig:architecture}). We investigated variations of recurrent architectures and hyper-parameters, and settled on a simple single layer LSTM, but most RNN's produces similar results.

The blue curve of Figure \ref{fig:accuracy-vs-slices} shows the results of adding a recurrent bottleneck on top of a spatially localized receptive field and a stateful NMS for both baseline architectures. Generically, we observe that predictive performance increases across the entire range of $n$. The performance gains are more notable across increasing number of slices $n$,
as well as all vehicle object sizes
(Table \ref{table:error-analysis}). These results are expected given that we observed systematic failure introduced for large objects that subtend multiple slices (Figure \ref{fig:accuracy-vs-slices}). For instance, with PointPillars at $n=32$ slices, the mAP for vehicles is boosted from 39.2\% to 48.9\% through the addition of a learned, recurrent architecture (Figure \ref{fig:accuracy-vs-slices}, blue curves). This streaming model is only marginally lower than the mAP of the original model that has access to the complete LiDAR scan.

Although a streaming model sacrifices slightly in terms of predictive performance, we expect that the resulting streaming model would realize substantial gains in terms of computational demand and latency. In the following section we explore this question in detail.

\subsection{Streaming detection reduces latency and peak computational demand}

\begin{figure}[t]
\begin{center}
\includegraphics[width=0.95\linewidth]{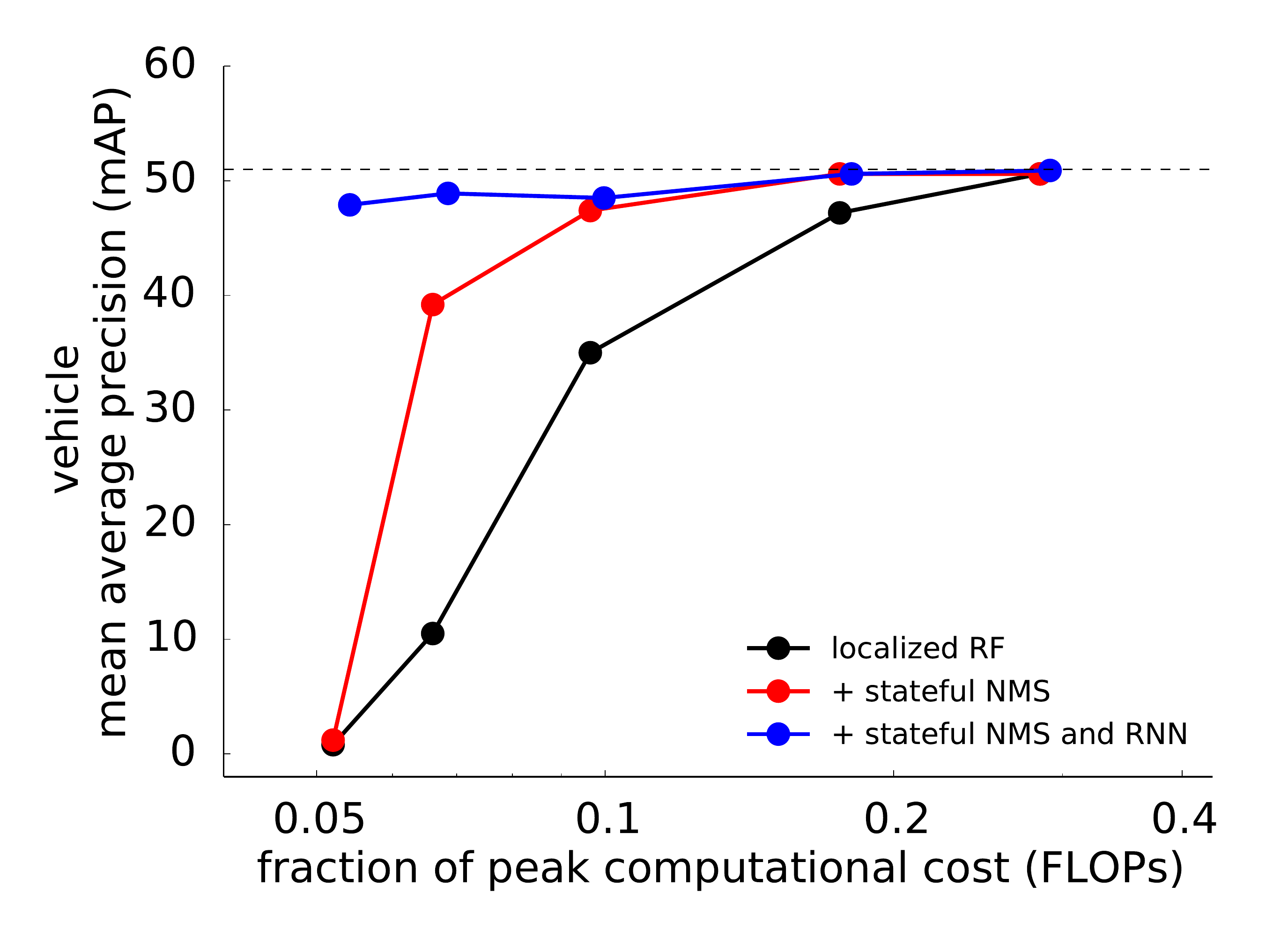}
\caption{Fraction of peak computational demand (FLOPs) versus detection accuracy (vehicle mAP) across varying number of slices $n$ for \cite{lang2018pointpillars}. Note the logarithmic scale on the x-axis. Each curve represents streaming architecture (see text). Dashed line indicates non-streaming baseline.}
\label{fig:accuracy-vs-flops}
\end{center}
\vspace{-0.5cm}
\end{figure}

\begin{figure}[t]
\begin{center}
\includegraphics[width=0.95\linewidth]{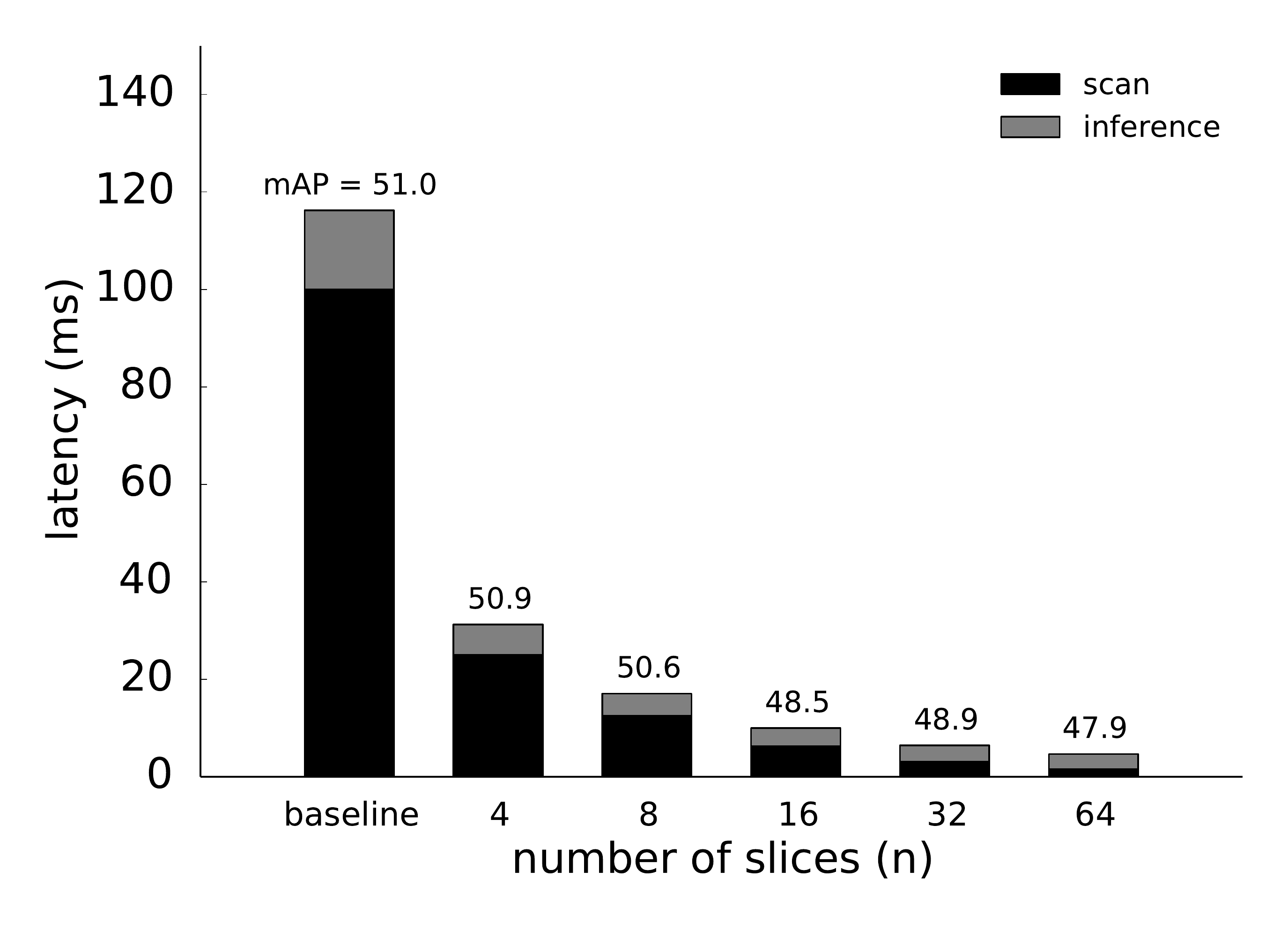}
\caption{Worst-case latency from the initial measurement of the vehicle to the detection for non-streaming (baseline) and streaming detection model (stateful NMS and RNN), broken down by phase. Scan phase latency is based on \mbox{10 Hz} LiDAR period \cite{waymo-open-dataset}. Inference phase latency estimated from baseline GPU implementation \cite{lang2018pointpillars}. Numbers on top of bar are vehicle mAP from the detection model.}
\label{fig:accuracy-vs-flops-2}
\end{center}
\end{figure}

The impetus for pursuing a streaming based model for detection on LiDAR is that we expect such a system to improve \textbf{end-to-end} latency for detecting objects. We define this latency as the duration of time between the earliest observation of an object (i.e. the earliest time at which reflecting LiDAR points are received) and the identification of a localized, labeled object.
In practice, for a non-streaming model the latency corresponds to the summation of the time for a complete (worst-case) $360^o$ rotation and the subsequent inference operation on the complete LiDAR scene. The latency in a streaming detection system should be improved for two reasons: (1) the computational demand for processing a fraction of the LiDAR spin should be roughly $\frac{1}{n}$ of the complete scene and (2) the inference operation does not require artificially waiting for the full LiDAR spin and may be pipelined. In this section, we focus our analysis on \cite{lang2018pointpillars} because of previously reported latencies.

To test the first point, we compute the theoretical peak computational demand (in FLOPS) for running single frame inference on the baseline model as well as streaming models. Figure \ref{fig:accuracy-vs-flops} compares the peak FLOPS versus the detection accuracy across varying slices $n$ for each of the 3 streaming architectures. We display the compute of each approach in terms of the \emph{fraction} of peak FLOPS required for the non-streaming baseline model (Note the log x-axis). We observe that a model with a localized receptive field (black curve) reveals a trade-off in accuracy versus the amount of computational demand. However, subsequent models with stateful NMS (red curve) and stateful NMS and RNN (blue curve) require much fewer peak FLOPS to achieve most of the detection performance of the baseline. Furthermore, the stateful NMS and RNN achieves nearly baseline predictive performance across a wide range of slices $n$ with a computational cost of roughly $\frac{1}{n}$. Thus, the streaming model requires less peak computational demand with minimal degradation in predictive performance.

Taking into account both the earlier triggering of inference and the reduced computational demand for each slice, we next investigate how streaming models can reduce end-to-end latency for detecting objects.
Unfortunately, latency is very heavily determined by the inference hardware, as well as the rotational period of the LiDAR system. To estimate reasonable speed up gains, we test these ideas on a previously reported implementation speed for the PointPillars model \cite{lang2018pointpillars} (Section 6), and employ the rotational period of the Waymo Open Dataset (i.e. 100 ms for 10 Hz) \cite{waymo-open-dataset}. Figure \ref{fig:accuracy-vs-flops-2} plots the latency versus the detection accuracy across the streaming model variants; we assume that the scan time is equivalent to the worst case delay between the first measurement of the object and the triggering of inference (the end of the slice): e.g., with 4 slices and 10Hz period, each slice triggers inference every 25ms.
We estimate inference latency by scaling the floating point computation time \cite{lang2018pointpillars} by the fraction of point cloud scene $\frac{1}{n}$ observed within an angular wedge necessary for inference.

The streaming models reduce end-to-end latency significantly in comparison to a non-streaming baseline model. In fact, we observe that for $n=8$, a streaming model with a stateful NMS and RNN achieves competitive detection accuracy with the non-streaming model, but with $\frac{1}{15}$ of the latency (17ms vs 116ms). We take these results to indicate as a proof of concept that a streaming detection system may significantly reduce end-to-end latency without significantly sacrificing predictive accuracy.

\subsection{Increasing model size exceeds baseline, maintaining favorable computational demand}

\begin{figure*}[t]
\begin{center}
{\bf PointPillars} \\
\includegraphics[width=0.39\linewidth]{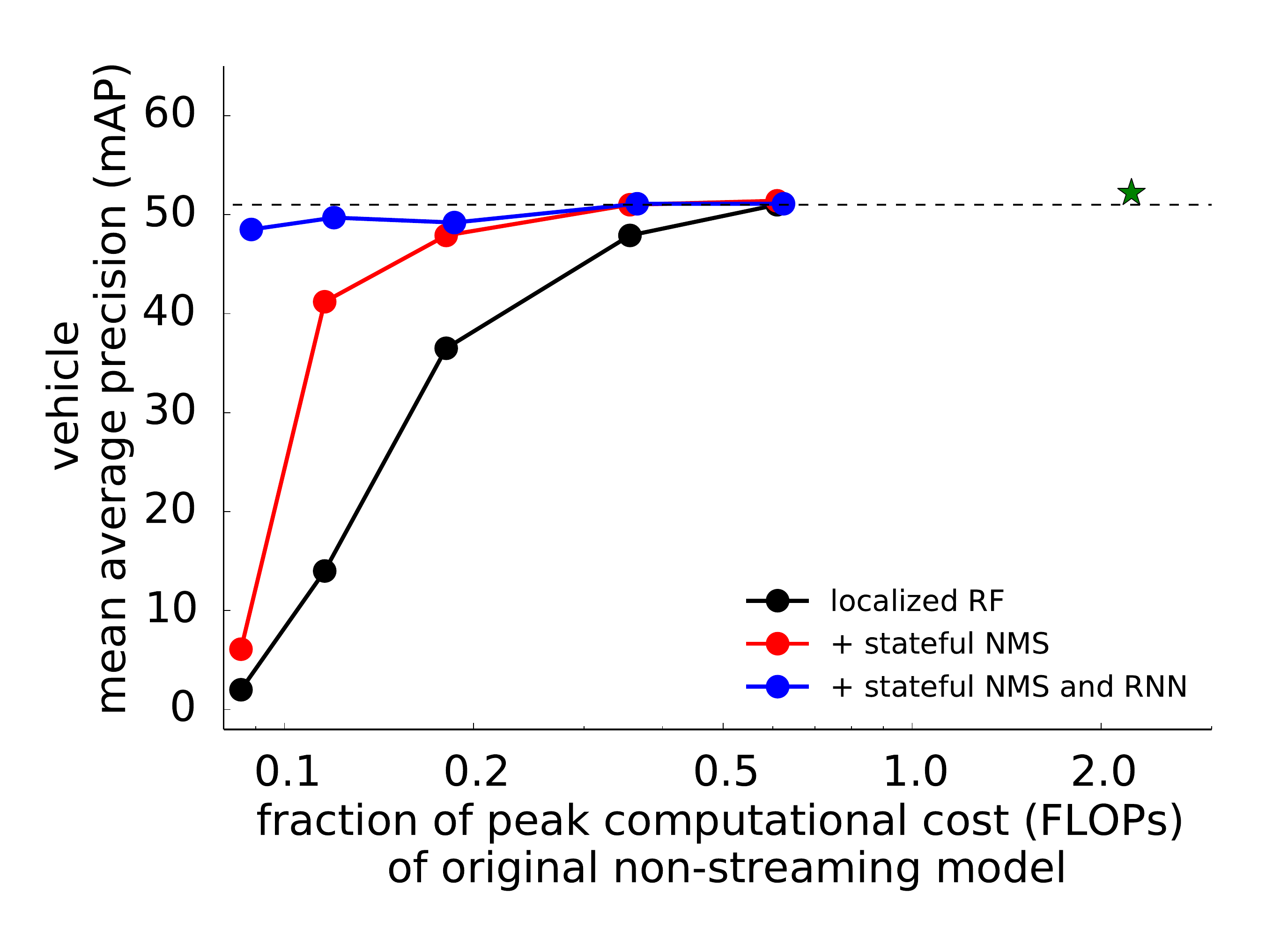}
\includegraphics[width=0.39\linewidth]{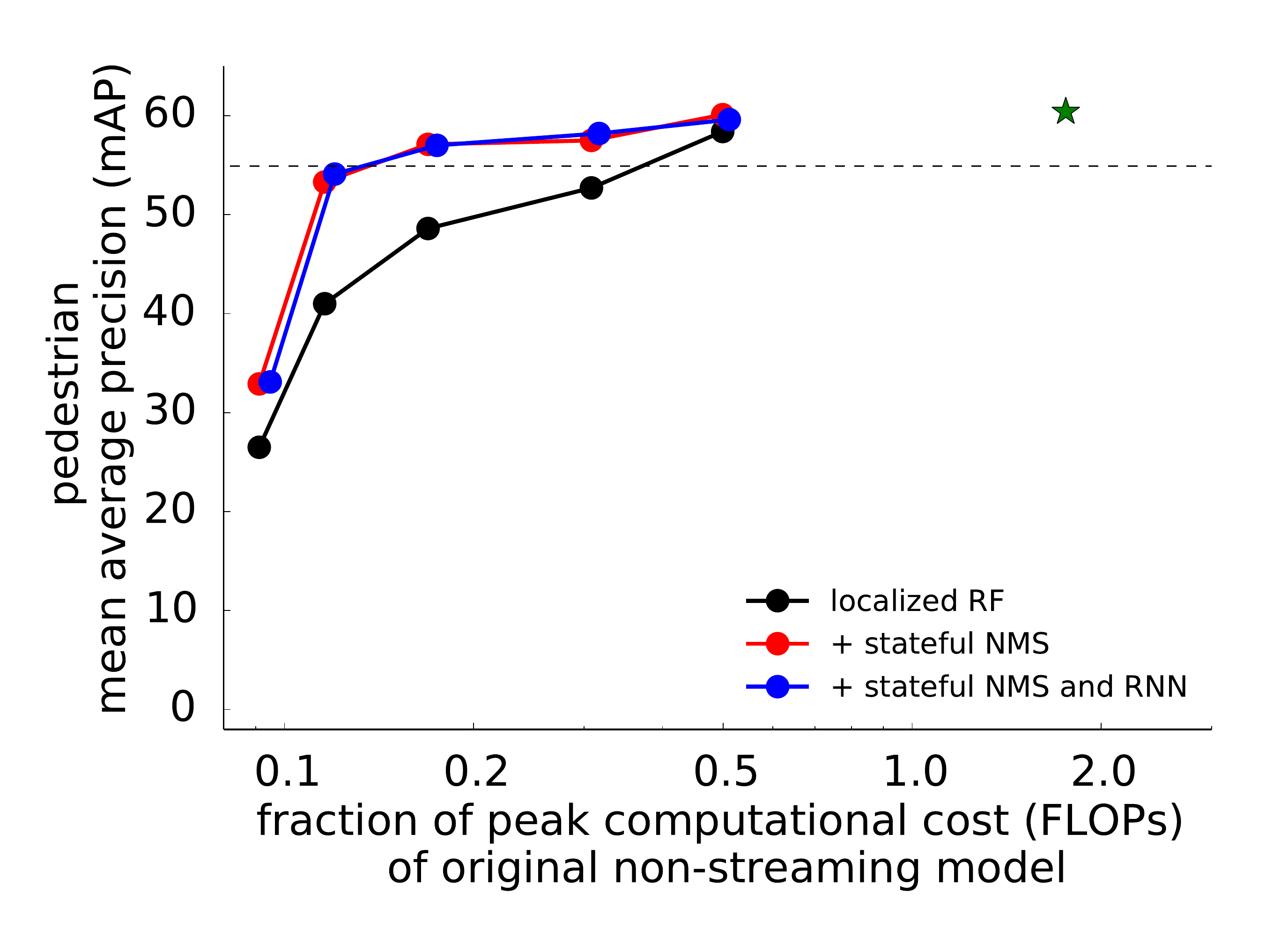} \\
{\bf StarNet} \\
\includegraphics[width=0.39\linewidth]{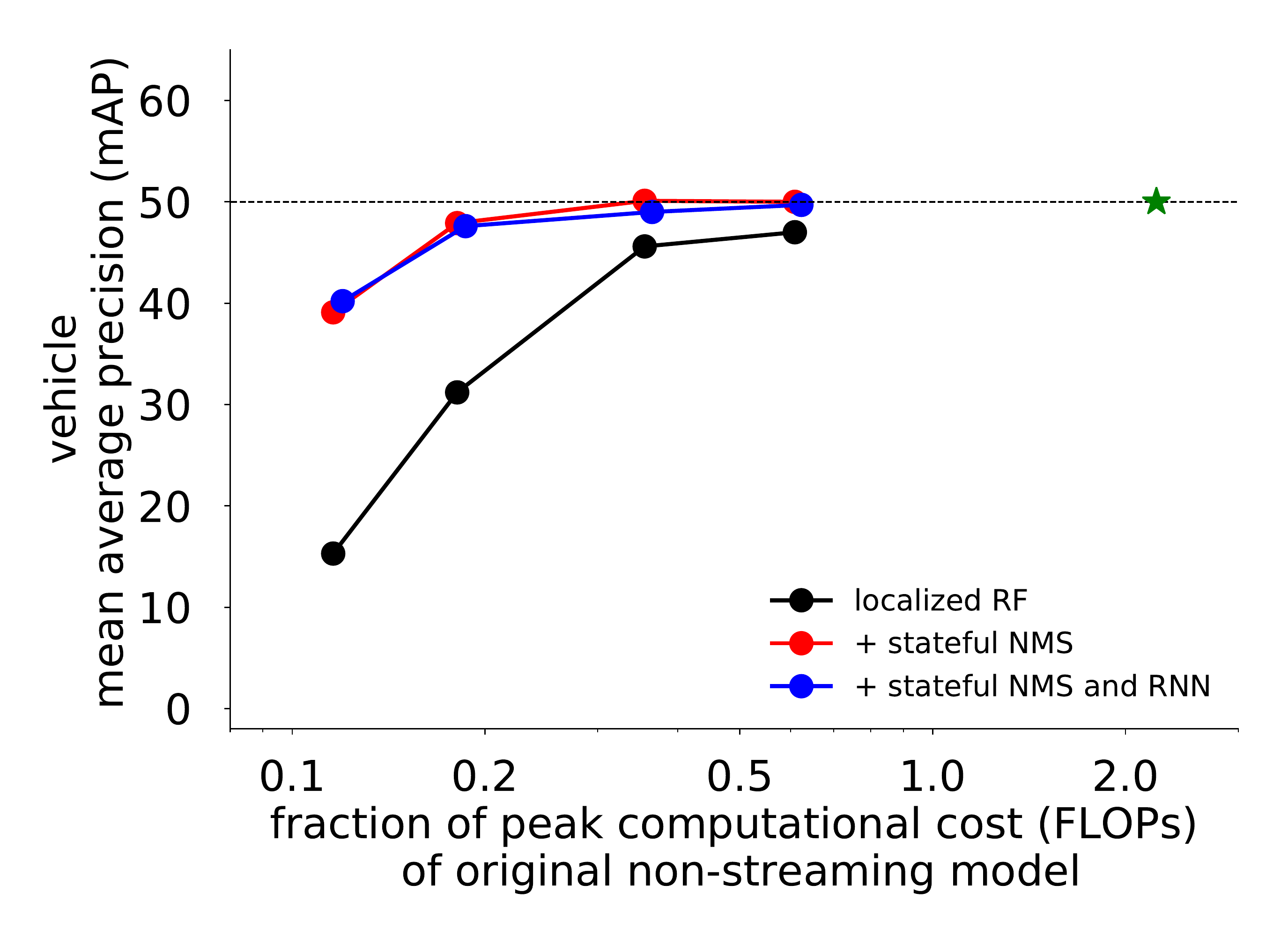}
\includegraphics[width=0.39\linewidth]{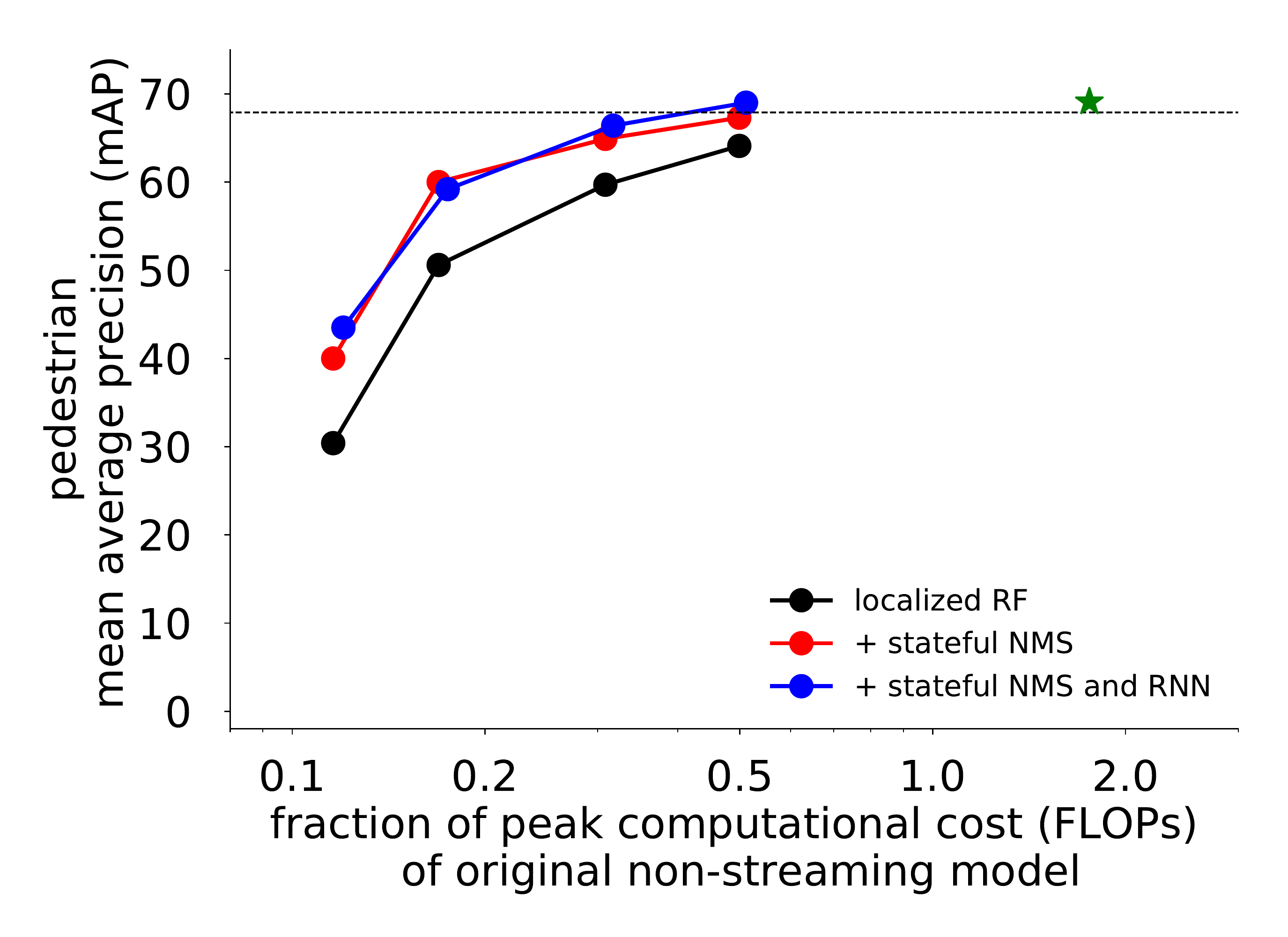}
\end{center}
\caption{{\bf A larger streaming model may exceed the baseline performance, but with a fraction of the peak computational budget.} Results presented for vehicle (left) and pedestrian (right) detection on
the larger (top) PointPillars \cite{lang2018pointpillars} and (bottom) StarNet \cite{ngiam2019starnet} relative to the original non-streaming baseline. Green star indicates the relative peak FLOPS and accuracy of the larger non-streaming model. Axes and legend follow Figure \ref{fig:accuracy-vs-flops}. Computational cost varies inversely with the number of slices($n$) from 4 to 64.}
\label{fig:exceeding-baseline}
\end{figure*}

The resulting streaming models achieve competitive predictive performance but with a fraction of the peak computational demand and latency. We next ask if one may further improve streaming model performance, while maintaining a decreased peak computational demand. It is well-known in the deep learning literature that increasing model size may lead to increased predictive performance (e.g. \cite{ciresan2010,Coates91ananalysis,lecun2015deep}).
Broadly speaking, we attempt to double the computational cost of each baseline network in order to attempt to boost the overall performance.
For instance, for PointPillars we increase the size of the feature pyramid by systematically increasing the spatial resolution of the activation map. Specifically, we increase the top-down view grid size from 384 to 512 for pedestrian models, and from 512 to 784 for vehicle models, resulting in models with 1.75$\times$ and 2.23$\times$ relative expense, respectively. For StarNet, we increase the size of all hidden unit representation by a factor of 1.44$\times$.

Figure \ref{fig:exceeding-baseline} shows the predictive accuracy of the resulting both streaming baseline models on vehicles (left) and pedestrians (right) across slices $n$ in comparison to the non-streaming baseline for both architectures.
As a point of reference, we show the relative peak computational cost of these larger models when run in a non-streaming node (green star). 
The x-axis measures logarithmically the peak computational demand of the resulting model expressed as a fraction of the non-streaming baseline model.  
Importantly, we observe that even though the peak computational demand is a small fraction of the non-streaming baseline model, the predictive performance matches or exceeds the non-streaming baseline model (e.g. 60.1 mAP versus 54.9 mAP for pedestrians with PointPillars).
Note that in order to achieve these gains in the non-streaming model requires increasing the peak computational cost by 2.25$\times$ (green star).
Moreover, at the lower peak FLOPS count, the stateful NMS and RNN model retains the most accuracy of the original baseline, yet may achieve $>3\times$ latency gains.
We take these results to indicate that much opportunity exists for further improving streaming models to well exceed a non-streaming model, while maintaining significantly reduced latency compared to a non-streaming architecture.

\section{Discussion}
In this work, we have described streaming object detection for point clouds for self-driving car perceptions systems.
Such a problem offers an opportunity for blending ideas from object detection \cite{ren2015faster,girshick2015fast,liu2016ssd,redmon2016you}, tracking \cite{liu2018mobile,liu2019looking,feichtenhofer2017detect} and sequential modeling \cite{wu2016google}.
Streaming object detection offers the opportunity to detect objects in a SDC environment that significantly minimizes latency (e.g. $3$-$15\times$) and better utilizes limited computational resources.

We find that simple methods based on restricting the receptive field, adding temporal state to the non-maximum suppression, and learning a perception state across time via recurrence suffice for providing competitive if not superior detection performance on a large-scale self-driving dataset.
The resulting system achieves favorable computational performance ($\sim 1/10^{th}$) and improved expected latency ($\sim 1/15^{th}$) with respect to a baseline non-streaming system. Such gains provide headroom to scale up the system to surpass baseline performance (60.1 vs 54.9 mAP) while maintaining a peak computational budget far below a non-streaming model.

This work offers opportunity for further improving this methodology, or application to new streaming sensors (e.g. high-resolution cameras that emit rasterized data in slices). While this work focuses on streaming models for a single frame, it is possible to also extend the models to incorporate data across multiple frames. 
We note that a streaming model may be amendable towards tracking problems since it already incorporates state. Finally, we have explored meta-architecture changes with respect to two competitive object detection baselines \cite{lang2018pointpillars,ngiam2019starnet}. 
We hope our work will encourage further research on other point cloud based perception systems to test their efficacy in a streaming setting \cite{yang2018pixor,luo2018fast,yang2018hdnet,lang2018pointpillars,meyer2019lasernet,zhou2018voxelnet,yan2018second,ngiam2019starnet}.

\section*{Acknowledgements}
\noindent We thank the larger teams at Google Brain and Waymo for their help and support. We also thank Chen Wu, Pieter-jan Kindermans, Matthieu Devin and Junhua Mao for detailed comments on the project and manuscript.

{\small
\bibliographystyle{style/ieee_fullname}
\bibliography{paper}
}

\clearpage
\appendix

\section*{Appendix}


\section{Architecture and Training Details}

\begin{table}[!h]
\centering
{\small
\begin{tabular}{rccccc} \toprule
Operation               & Stride    & \# In & \# Out    & Activation  & Other  \\
\midrule
{\bf Streaming detector} \\
Featurizer MLP          &           & $12$              & $64$                  &               & With max pooling  \\
Convolution block   & $1$       & $64$              & $64$                  & ReLU          & Layers=$4$        \\
Convolution block   & $2$       & $64$              & $128$                 & ReLU          & Layers=$6$        \\
Convolution block   & $2$       & $128$             & $256$                 & ReLU          & Layers=$6$        \\
LSTM                    &           & $256$             & $256$                 &               & Hidden=$128$      \\
Deconvolution $1$       & $1$       & $64$              & $128$                 & ReLU          &                   \\
Deconvolution $2$       & $2$       & $128$             & $128$                 & ReLU          &                   \\
Deconvolution $3$       & $4$       & $256$             & $128$                 & ReLU          &                   \\
Convolution/Detector    & $1$       & $384$             & $16$                  &               & Kernel=$3\times 3$\\ \midrule
{\bf Convolution block} \\
$(S, C_{in}, C_{out}, L)$                                                                \\
Convolution $1$         & $S$       & $C_{in}$          & $C_{out}$             & ReLU          & Kernel=$3\times 3$\\
Convolution $2, ..., L-1$ & $1$     & $C_{out}$         & $C_{out}$             & ReLU          & Kernel=$3\times 3$\\ \midrule
Normalization          & \multicolumn{5}{l}{Batch normalization before ReLU for every} \\
& \multicolumn{5}{l}{convolution and deconvolution layer} \\
Optimizer              & \multicolumn{5}{l}{Adam \cite{kingma2014adam} ($\alpha = 0.001$, $\beta_1 = 0.9$, $\beta_2 = 0.999$)}  \\
Parameter updates      & \multicolumn{5}{l}{40,000 - 80,000}                     \\
Batch size             & \multicolumn{5}{l}{64}                         \\
Weight initialization  & \multicolumn{5}{l}{Xavier-Glorot\cite{glorot2010understanding}}  \\ \bottomrule
\end{tabular}
}
\caption{PointPillars detection baseline \cite{lang2018pointpillars}.}
\label{table:pointpillars-details}
\end{table}
\vspace{-0.3cm}


\begin{table}[!h]
\centering
{\small
\begin{tabular}{rcccc} \toprule
Operation               & \# In & \# Out    & Activation  & Other  \\
\midrule
{\bf Streaming detector} \\
Linear                  & $4$              & $64$    &  ReLU & \\
StarNet block $\times$ $5$                  & $64$              & $64$    &  ReLU & Final feature is the concat of all layers\\
LSTM                      & $384$             & $384$                 &               & Hidden=$128$      \\
Detector           & $384$             & $16$                  &               & \\ \midrule
{\bf StarNet block} \\
Max-Concat     & $64$          & $128$             &     & \\
Linear     & $128$          & $256$             &  ReLU   & \\
Linear     & $256$          & $64$             &    ReLU & \\
 \midrule
Normalization          & \multicolumn{4}{l}{Batch normalization before ReLU} \\
Optimizer              & \multicolumn{4}{l}{Adam \cite{kingma2014adam} ($\alpha = 0.001$, $\beta_1 = 0.9$, $\beta_2 = 0.999$)}  \\
Parameter updates      & \multicolumn{4}{l}{100,000}                     \\
Batch size             & \multicolumn{4}{l}{64}                         \\
Weight initialization  & \multicolumn{4}{l}{Kaiming Uniform \cite{he2015delving}}  \\ \bottomrule
\end{tabular}
}
\caption{StarNet detection baseline \cite{ngiam2019starnet}.}
\label{table:starnet-details}
\end{table}

\end{document}